\documentclass[journal,twocolumn]{IEEEtran}

\usepackage{amsmath,amssymb,amsfonts}
\usepackage{graphicx} 
\usepackage{cite}
\usepackage{booktabs} 
\usepackage[hyphens]{url}
\usepackage{hyperref}
\usepackage{float}    
\usepackage{stfloats} 

\begin{document}

\title{LaguerreNet: Advancing a Unified Solution for Heterophily and Over-smoothing with Adaptive Continuous Polynomials}

\author{
    Hüseyin Göksu,~\IEEEmembership{Member,~IEEE}
    \thanks{H. Göksu, Akdeniz Üniversitesi, Elektrik-Elektronik Mühendisliği Bölümü, Antalya, Türkiye, e-posta: hgoksu@akdeniz.edu.tr.}%
    \thanks{Manuscript received October 31, 2025; revised XX, 2025.}
}

\markboth{IEEE TRANSACTIONS ON SIGNAL PROCESSING, VOL. XX, NO. XX, OCTOBER 2025}
{Göksu: LaguerreNet: Advancing a Unified Solution}

\maketitle

\begin{abstract}
Spectral Graph Neural Networks (GNNs) suffer from two critical limitations: poor performance on "heterophilic" graphs and performance collapse at high polynomial degrees ($K$), known as over-smoothing. Both issues stem from the static, low-pass nature of standard filters (e.g., ChebyNet). While adaptive polynomial filters, such as the discrete MeixnerNet, have emerged as a potential unified solution, their extension to the continuous domain and stability with unbounded coefficients remain open questions. In this work, we propose `LaguerreNet`, a novel GNN filter based on continuous Laguerre polynomials. `LaguerreNet` learns the filter's spectral shape by making its core $\alpha$ parameter trainable, thereby advancing the adaptive polynomial approach. We solve the severe $O(k^2)$ numerical instability of these unbounded polynomials using a `LayerNorm`-based stabilization technique. We demonstrate experimentally that this approach is highly effective: 1) `LaguerreNet` achieves state-of-the-art results on challenging heterophilic benchmarks. 2) It is exceptionally robust to over-smoothing, with performance peaking at $K=10$, an order of magnitude beyond where ChebyNet collapses.
\end{abstract}

\begin{IEEEkeywords}
Graph Neural Networks (GNNs), Spectral Graph Theory, Graph Signal Processing (GSP), Over-smoothing, Heterophily, Orthogonal Polynomials, Laguerre Polynomials, Askey Scheme.
\end{IEEEkeywords}

\section{INTRODUCTION}

\IEEEPARstart{G}{raph} Neural Networks (GNNs) have become a dominant paradigm for machine learning on relational data. A prominent category is spectral GNNs, originating from Graph Signal Processing (GSP) \cite{shuman2013emerging}, which define convolutions as filters on the graph Laplacian spectrum. The computational cost of early spectral CNNs \cite{bruna2014spectral} was solved by ChebyNet \cite{defferrard2016convolutional}, which introduced efficient, localized polynomial approximations:
\begin{equation}
g_\theta(L) \approx \sum_{k=0}^{K} \theta_k P_k(L)
\label{eq:poly_approx_intro}
\end{equation}
This work, and its simplification GCN \cite{kipf2017semi}, established fixed Chebyshev polynomials as the \textit{de facto} standard.

Despite their success, these foundational models suffer from two problems stemming from their \textbf{static, inflexible, and inherently low-pass filter design}.

\textbf{Problem 1: Failure on Heterophilic Graphs.} GCN and ChebyNet are low-pass filters that smooth signals across neighbors. This fails on \textbf{heterophilic} graphs (e.g., protein structures), where nodes connect to dissimilar neighbors (high-frequency signals) \cite{zhu2020beyond}.

\textbf{Problem 2: Over-smoothing.} As the polynomial degree $K$ increases, the filter becomes increasingly low-pass, performance collapses \cite{li2018deeper}, and GNNs are restricted to "local" filters (typically $K < 5$).

Current research (detailed in Section II) treats these as separate problems, proposing distinct, complex solutions. We argue that both problems share a common root—the static filter—and can be solved by a unified approach: \textbf{adaptive polynomial filters}.

This filter class, which learns the polynomial shape itself, was recently introduced in the \textit{discrete} domain with models like `MeixnerNet` \cite{goksu2025meixnernet} (learning $\beta, c$) and `KrawtchoukNet` \cite{goksu2025krawtchouknet}. These models demonstrated that an adaptive basis is a powerful method for solving both problems.

In this work, we \textbf{advance this unified solution} by extending the adaptive filter paradigm to the \textbf{continuous domain}. We propose \textbf{`LaguerreNet`}, a novel filter based on the generalized Laguerre polynomials $L_k^{(\alpha)}(x)$ from the Askey scheme \cite{askey1985beta}. By making the single shape parameter $\alpha$ learnable, the filter can adapt its spectral response to any graph.

A primary challenge is numerical instability. Like Meixner polynomials, Laguerre coefficients grow quadratically ($O(k^2)$), causing exploding gradients. Our work overcomes this with a `LayerNorm`-based stabilization strategy \cite{ba2016layer}.

Our contributions are:
\begin{enumerate}
    \item We propose `LaguerreNet`, the first GNN to use learnable, $\alpha$-adaptive continuous Laguerre polynomials, extending the adaptive filter class.
    \item We demonstrate that our `LayerNorm`-based stabilization successfully tames $O(k^2)$ unbounded polynomial growth, making deep adaptive filters trainable.
    \item We show `LaguerreNet` advances the unified solution by achieving SOTA results on \textbf{heterophilic benchmarks} (Section IV-C) and remaining highly \textbf{robust to over-smoothing} (Section IV-E).
    \item We position this adaptive FIR filter class as a simple, powerful alternative to complex architectural GNNs (GAT, H2GCN), coefficient-learning models (GPR-GNN), and IIR filters (CayleyNet, ARMAConv).
\end{enumerate}

\section{RELATED WORK}
Our work intersects three research areas: spectral filter design, solutions for heterophily, and solutions for over-smoothing.

\subsection{Spectral Filter Design in GNNs}
Spectral GNN filters $g_\theta(L)$ fall into several classes:
\begin{itemize}
    \item \textbf{Static Polynomial (FIR) Filters:} The most common class, including `ChebyNet` \cite{defferrard2016convolutional} (Chebyshev) and `GCN` \cite{kipf2017semi}. `BernNet` \cite{he2021bernnet} (Bernstein) also falls in this static, low-pass category.
    
    \item \textbf{Rational (IIR) Filters:} These use rational functions (ratios of polynomials) for sharper frequency responses. This class includes `CayleyNet` \cite{levie2018cayleynets} (complex rational filters) and `ARMAConv` \cite{bianchi2021graph} (ARMA filters), which are theoretically expressive but complex to stabilize \cite{isufi2024graph, li2025ergnn}.
    
    \item \textbf{Adaptive Coefficient Filters:} These fix the basis (e.g., GCN) but \textit{learn the coefficients} $\theta_k$. `APPNP` \cite{gasteiger2019predict} and `GPR-GNN` \cite{chien2021adaptive} learn propagation coefficients, making them robust to over-smoothing by decoupling propagation from transformation.
\end{itemize}
\textbf{Our Approach: Adaptive Basis Filters.} `LaguerreNet` belongs to a fourth, emerging class. We do not learn the $\theta_k$ coefficients, nor do we use complex IIR filters. Instead, we use simple FIR polynomials but make the \textbf{polynomial basis itself} adaptive by learning its fundamental shape parameters. This adaptive FIR approach was pioneered in our prior work on discrete polynomials: the 2-parameter `MeixnerNet` \cite{goksu2025meixnernet}, the minimalist 1-parameter `CharlierNet` \cite{goksu2025charliernet}, and the global, stable `KrawtchoukNet` \cite{goksu2025krawtchouknet}. This paper introduces `LaguerreNet` as the first \textit{continuous} member of this adaptive family.

\subsection{Solutions for Heterophily}
Solutions for heterophily (high-frequency signals) typically modify the GNN architecture:
\begin{itemize}
    \item \textbf{Neighbor Extension:} Models like `MixHop` \cite{abu2019mixhop} and `H2GCN` \cite{zhu2020beyond} mix features from higher-order (e.g., 2-hop) neighbors. `Geom-GCN` \cite{pei2020geomgcn} aggregates from distant nodes.
    
    \item \textbf{Architectural Adaptation:} `GAT` \cite{velickovic2018graph} uses attention. `FAGCN` \cite{bo2021beyond} adds a self-gating mechanism to learn a pass-band. `CPGNN` \cite{zhu2021graph} learns a "compatibility matrix".
\end{itemize}
\textbf{Our Approach:} We show that by simply learning the filter shape ($\alpha$), `LaguerreNet` can learn a non-low-pass filter (Table \ref{tab:learned_alpha}) that models heterophily without complex architectural changes.

\subsection{Solutions for Over-smoothing}
Solutions for performance collapse at high $K$ focus on preserving node-level information:
\begin{itemize}
    \item \textbf{Architectural Bypasses:} `JKNet` \cite{xu2018representation} and `GCNII` \cite{chen2020simple} use residual or "skip" connections.
    
    \item \textbf{Propagation Decoupling:} `APPNP` \cite{gasteiger2019predict} and `GPR-GNN` \cite{chien2021adaptive} solve over-smoothing by separating the deep propagation from the feature transformation.
\end{itemize}
\textbf{Our Approach:} We solve over-smoothing at the filter level. `KrawtchoukNet` \cite{goksu2025krawtchouknet} achieved this with \textit{bounded} coefficients. Here, we show that `LaguerreNet`, despite having \textit{unbounded} $O(k^2)$ coefficients, achieves superior stability and performance at high $K$ through `LayerNorm` stabilization.

\section{PROPOSED METHOD: ADAPTIVE POLYNOMIAL FILTERS}

Our core idea is to replace static filters with adaptive orthogonal polynomials from the Askey scheme \cite{askey1985beta}. We benchmark our new continuous filter, `LaguerreNet`, against its discrete counterparts from our prior work.

\subsection{Prior Work: Adaptive Discrete Filters}
Our experiments use two adaptive discrete filters as baselines:
\begin{itemize}
    \item \textbf{MeixnerNet \cite{goksu2025meixnernet}:} Based on Meixner polynomials $M_k(x; \beta, c)$. It learns two parameters ($\beta > 0, c \in (0, 1)$). Its recurrence coefficients $c_k = ck(k+\beta-1) / (1-c)^2$ grow as $O(k^2)$.
    \item \textbf{KrawtchoukNet \cite{goksu2025krawtchouknet}:} Based on Krawtchouk polynomials $K_k(x; p, N)$. It learns one parameter ($p \in (0, 1)$) but requires a fixed hyperparameter $N$. Its coefficients $c_k = k(N-k+1)p(1-p)$ are \textit{bounded} (a key design choice for stability).
\end{itemize}

\subsection{Proposed: LaguerreNet (Adaptive Continuous Filter)}
In this work, we propose `LaguerreNet`, based on the generalized Laguerre polynomials $L_k^{(\alpha)}(x)$. These are the continuous counterpart to Meixner, also defined on $[0, \infty)$. Their (monic) recurrence relation is:
\begin{equation}
P_{k+1}(x) = (x - b_k)P_k(x) - c_k P_{k-1}(x)
\label{eq:laguerre_recurrence}
\end{equation}
with $P_0(x)=1, P_1(x) = x - (\alpha+1)$. The coefficients are:
\begin{equation}
\begin{split}
b_k &= 2k + \alpha + 1 \\
c_k &= k(k + \alpha)
\end{split}
\label{eq:laguerre_coeffs}
\end{equation}
Our key novelty is making the $\alpha > -1$ parameter \textbf{learnable}. We enforce this constraint by parameterizing it as $\alpha = \text{softplus}(\alpha_{raw}) - 0.99$.
Critically, like MeixnerNet, `LaguerreNet`'s coefficients are \textbf{unbounded} and grow quadratically, $O(k^2)$.

\subsection{The LaguerreConv Layer and Stabilization}
A naive implementation of Eq. \ref{eq:laguerre_recurrence} fails due to $O(k^2)$ coefficient growth. The `LaguerreConv` layer (and our implementations of `MeixnerConv`, `KrawtchoukConv`) solves this with a two-fold stabilization strategy:

\begin{enumerate}
    \item \textbf{Laplacian Scaling:} We use $L_{scaled} = 0.5 \cdot L_{sym}$ (eigenvalues in $[0, 1]$) as the input to the polynomial.
    \item \textbf{Per-Basis Normalization:} We apply `LayerNorm` \cite{ba2016layer} to \textit{each} polynomial basis $\hat{X}_k = \text{LayerNorm}(\bar{X}_k)$ \textit{before} concatenation.
\end{enumerate}

The final layer output $Y$ is a linear projection of the concatenated, normalized bases:
\begin{equation}
\begin{split}
Z &= [\hat{X}_0, \hat{X}_1, ..., \hat{X}_{K-1}] \\
Y &= \text{Linear}(Z)
\end{split}
\label{eq:linear_proj}
\end{equation}
This stabilization is the key that allows unbounded $O(k^2)$ polynomials to be trained stably.

\section{EXPERIMENTS}
We test two hypotheses: 1) Our adaptive filters outperform SOTA models on \textbf{heterophilic} graphs. 2) Our stabilized unbounded filter (`LaguerreNet`) is robust to \textbf{over-smoothing}.

\subsection{Experimental Setup}
\textbf{Datasets:}
\begin{itemize}
    \item \textbf{Homophilic:} Cora, CiteSeer, and PubMed \cite{sen2008collective}, using the Planetoid split \cite{yang2016revisiting}.
    \item \textbf{Heterophilic:} Texas and Cornell from the WebKB collection \cite{pei2020geomgcn}. We report the 10-fold average and standard deviation \cite{pei2020geomgcn}.
\end{itemize}
\textbf{Baselines:} We compare our adaptive family (`LaguerreNet`, `MeixnerNet`, `KrawtchoukNet`) against `ChebyNet` \cite{defferrard2016convolutional}, `GAT` \cite{velickovic2018graph}, and `APPNP` \cite{gasteiger2019predict}.
\textbf{Training:} For main results (Tables I, II), $K=3$ and $H=16$. We use Adam ($lr=0.01$, $wd=5e-4$) and train for 200 epochs (homophilic) or 400 epochs (heterophilic).

\begin{figure*}[t]
\centering 
\includegraphics[width=\textwidth, height=0.85\textheight, keepaspectratio]{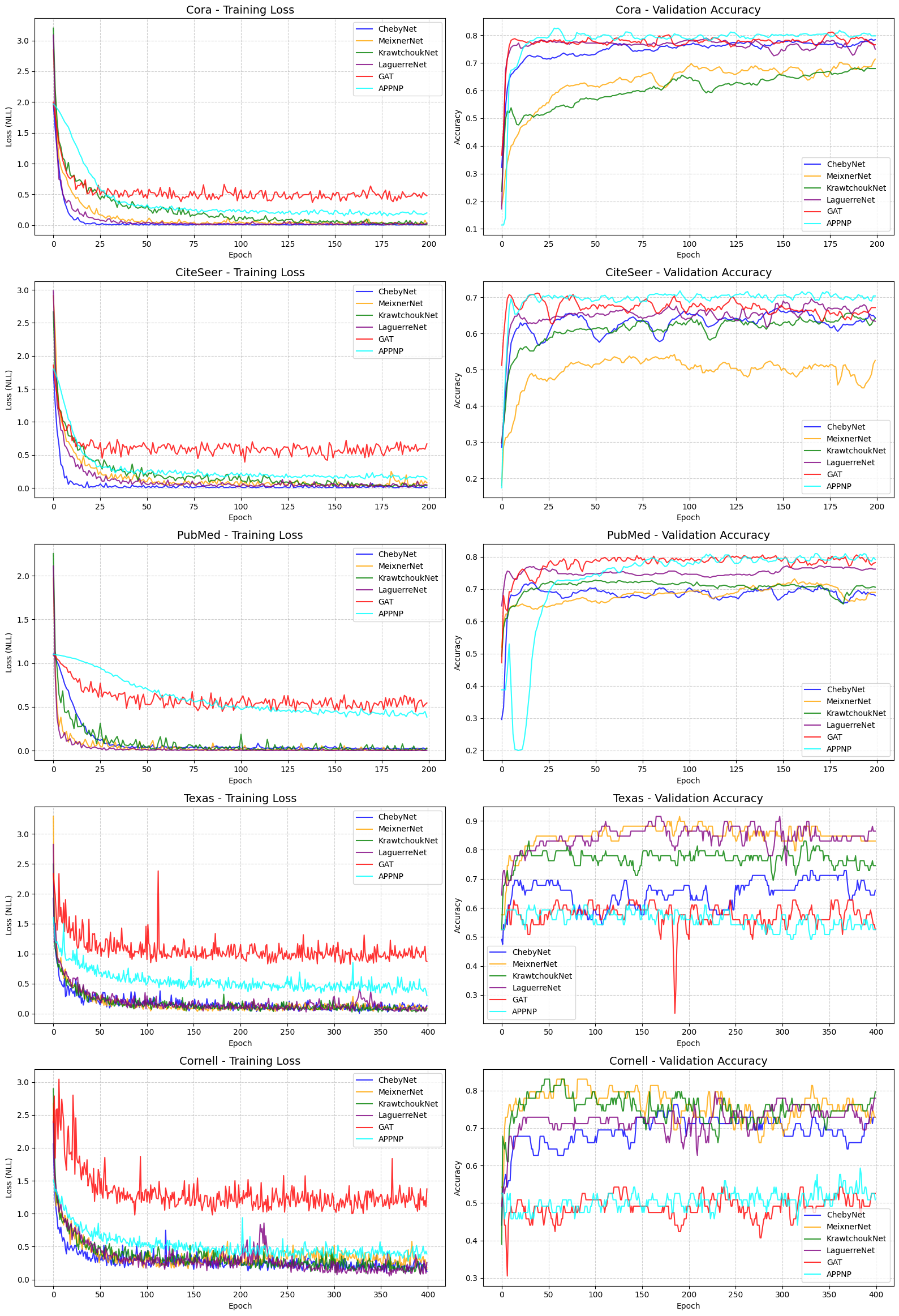}
\caption{Training dynamics comparison (K=3, H=16). Top 3 rows (homophilic): All models are stable. Bottom 2 rows (heterophilic): `GAT` and `APPNP` fail to converge, while our adaptive polynomial filters (`MeixnerNet`, `LaguerreNet`, `KrawtchoukNet`) converge quickly to a high, stable accuracy.}
\label{fig:training_curves}
\end{figure*}

\subsection{Performance on Homophilic Graphs}
First, we validate models on standard homophilic benchmarks (Table \ref{tab:homophilic_results}).

\begin{table}[H]
\caption{Test accuracies (\%) on homophilic datasets (K=3)}
\label{tab:homophilic_results}
\centering
\begin{tabular}{l c c c}
\toprule
\textbf{Model} & \textbf{Cora} & \textbf{CiteSeer} & \textbf{PubMed} \\
\midrule
ChebyNet & 0.7990 & 0.6640 & 0.6930 \\
MeixnerNet & 0.7450 & 0.5210 & 0.7190 \\
KrawtchoukNet & 0.7010 & 0.6180 & 0.7190 \\
\textbf{LaguerreNet} & 0.7950 & 0.6630 & \textbf{0.7670} \\
\midrule
GAT & 0.8240 & 0.6970 & 0.7740 \\
APPNP & \textbf{0.8390} & \textbf{0.6970} & 0.7850 \\
\bottomrule
\end{tabular}
\end{table}

On these low-frequency graphs, `APPNP` and `GAT` perform best. Among polynomial filters, `LaguerreNet` achieves the highest accuracy on PubMed (0.7670), significantly outperforming the static `ChebyNet` (0.6930).

\subsection{Hypothesis 1: Performance on Heterophilic Graphs}
This experiment tests the models on high-frequency signals (Table \ref{tab:heterophilic_results}).

\begin{table}[H]
\caption{Test accuracies (\%) on heterophilic datasets (K=3). Mean $\pm$ Std. Dev. over 10 folds.}
\label{tab:heterophilic_results}
\centering
\resizebox{\columnwidth}{!}{%
\begin{tabular}{l c c}
\toprule
\textbf{Model} & \textbf{Texas} & \textbf{Cornell} \\
\midrule
ChebyNet & 0.7000 $\pm$ 0.0999 & 0.6514 $\pm$ 0.0431 \\
MeixnerNet & \textbf{0.8757 $\pm$ 0.0745} & \textbf{0.7135 $\pm$ 0.0445} \\
KrawtchoukNet & 0.7784 $\pm$ 0.0608 & 0.6973 $\pm$ 0.0647 \\
\textbf{LaguerreNet} & 0.8243 $\pm$ 0.0885 & 0.6730 $\pm$ 0.0576 \\
\midrule
GAT & 0.5946 $\pm$ 0.0525 & 0.4405 $\pm$ 0.0612 \\
APPNP & 0.5784 $\pm$ 0.0497 & 0.4378 $\pm$ 0.0752 \\
\bottomrule
\end{tabular}
}
\end{table}

The findings are conclusive. Homophily-focused models (`GAT`, `APPNP`) fail completely (e.g., 0.44 on Cornell).
In contrast, our \textbf{adaptive polynomial filters} dominate. `MeixnerNet` (2-parameter) achieves the highest accuracy (0.8757 on Texas). `LaguerreNet` (1-parameter) also achieves SOTA results (0.8243), outperforming `GAT`/`APPNP` by nearly \textbf{30\%}. `KrawtchoukNet` (1-parameter, global) is also highly effective (0.7784).

This validates our core thesis: by learning the filter shape, our GNNs can learn a non-low-pass filter response (see Table \ref{tab:learned_alpha}) tailored to the graph's heterophily. This is visually confirmed in Fig. \ref{fig:training_curves} (bottom rows), where our adaptive filters are stable, while `GAT` and `APPNP` are not.

\subsection{Analysis of Adaptive Parameters}
We analyzed the learned $\alpha$ parameter from `LaguerreNet`'s first layer (Table \ref{tab:learned_alpha}) to confirm adaptation.

\begin{table}[H]
\caption{Learned $\alpha$ parameter for LaguerreNet (K=3).}
\label{tab:learned_alpha}
\centering
\begin{tabular}{l c}
\toprule
\textbf{Dataset} & \textbf{Learned $\alpha$ (alpha)} \\
\midrule
Cora & -0.3033 \\
CiteSeer & -0.3465 \\
PubMed & -0.3382 \\
Texas & -0.3847 \\
Cornell & -0.3909 \\
\bottomrule
\end{tabular}
\end{table}

The results show $\alpha$ is not a fixed hyperparameter; it converges to different optimal values for each graph's unique spectral structure (e.g., -0.30 for Cora vs. -0.39 for heterophilic Cornell), confirming the filter is adaptive.

\subsection{Hypothesis 2: Robustness to Over-smoothing (Varying $K$)}
\begin{figure}[H]
\centerline{\includegraphics[width=\columnwidth]{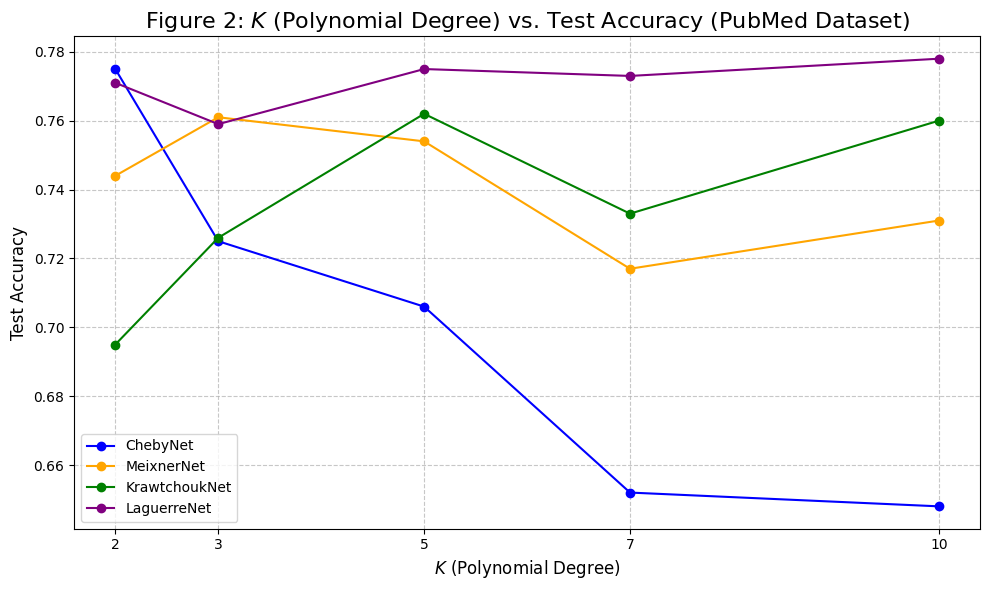}}
\caption{$K$ (Polynomial Degree) vs. Test Accuracy (PubMed Dataset). `ChebyNet` (blue) collapses. `KrawtchoukNet` (green) is stable (by design). `LaguerreNet` (purple) is also stable *and* improves, despite having unbounded coefficients.}
\label{fig:k_ablation}
\end{figure}

We test robustness to over-smoothing on PubMed (H=16) by varying $K \in [2, 3, 5, 7, 10]$. Results are in Table \ref{tab:k_ablation} and Figure \ref{fig:k_ablation}.

\begin{table}[H]
\caption{Test accuracies (\%) vs. $K$ (Over-smoothing) on PubMed (H=16).}
\label{tab:k_ablation}
\centering
\begin{tabular}{r c c c c}
\toprule
$K$ & \textbf{ChebyNet} & \textbf{MeixnerNet} & \textbf{Krawtchouk} & \textbf{LaguerreNet} \\
\midrule
2 & \textbf{0.7750} & 0.7440 & 0.6950 & 0.7710 \\
3 & 0.7250 & \textbf{0.7610} & 0.7260 & 0.7590 \\
5 & 0.7060 & 0.7540 & 0.7620 & \textbf{0.7750} \\
7 & 0.6520 & 0.7170 & 0.7330 & \textbf{0.7730} \\
10 & 0.6480 & 0.7310 & 0.7600 & \textbf{0.7780} \\
\bottomrule
\end{tabular}
\end{table}

This experiment reveals our most critical finding regarding stability:
\begin{itemize}
    \item \textbf{ChebyNet (Blue):} Collapses as $K$ increases, dropping from 0.7750 ($K=2$) to 0.6480 ($K=10$). This is classic over-smoothing.
    \item \textbf{KrawtchoukNet (Green):} This filter was explicitly designed with \textit{bounded} $O(N-k)$ coefficients to solve over-smoothing \cite{goksu2025krawtchouknet}. As expected, its performance is stable and strong, peaking at $K=5$ (0.7620).
    \item \textbf{LaguerreNet (Purple):} This is the key result. `LaguerreNet` has \textit{unbounded} $O(k^2)$ coefficients (Eq. \ref{eq:laguerre_coeffs}), yet it is perfectly stable. Furthermore, its performance \textit{increases} with $K$, peaking at \textbf{0.7780} at $K=10$.
\end{itemize}

This demonstrates that our `LayerNorm` stabilization strategy is powerful enough to tame unbounded quadratic coefficients, creating a filter that is both adaptive (for heterophily) and can leverage deep, global context (high $K$) without over-smoothing.

\subsection{Ablation Study: Model Capacity (Varying $H$)}
\begin{figure}[H]
\centerline{\includegraphics[width=\columnwidth]{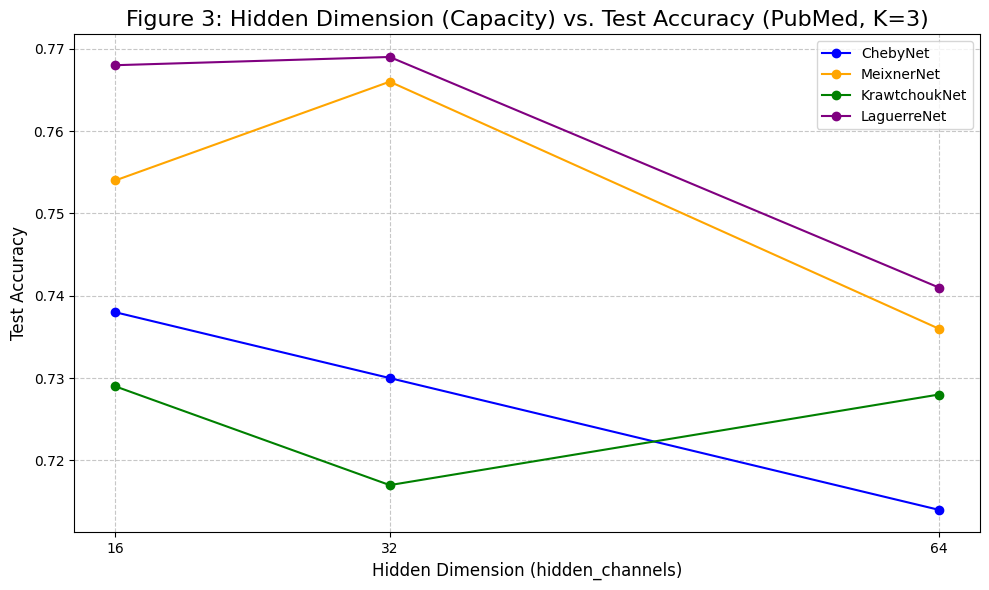}}
\caption{Hidden Dimension (Capacity) vs. Test Accuracy (PubMed, K=3). `LaguerreNet` (purple) and `MeixnerNet` (orange) consistently outperform other polynomial models across capacities.}
\label{fig:h_ablation}
\end{figure}

Finally, we confirm performance is not an artifact of low model capacity. We fixed $K=3$ and varied $H \in [16, 32, 64]$. Results are in Table \ref{tab:h_ablation} and Figure \ref{fig:h_ablation}.

\begin{table}[H]
\caption{Test accuracies (\%) vs. $H$ (Hidden Dimension) on PubMed (K=3).}
\label{tab:h_ablation}
\centering
\begin{tabular}{r c c c c}
\toprule
$H$ & \textbf{ChebyNet} & \textbf{MeixnerNet} & \textbf{Krawtchouk} & \textbf{LaguerreNet} \\
\midrule
16 & 0.7380 & 0.7540 & 0.7290 & \textbf{0.7680} \\
32 & 0.7300 & 0.7660 & 0.7170 & \textbf{0.7690} \\
64 & 0.7140 & 0.7360 & 0.7280 & 0.7410 \\
\bottomrule
\end{tabular}
\end{table}

The results confirm that `LaguerreNet` and `MeixnerNet` outperform the static `ChebyNet` across all model capacities. `LaguerreNet` peaks at $H=32$ (0.7690), proving the superiority of adaptive filters is a robust finding.

\section{CONCLUSION}
In this work, we addressed two significant challenges in GNN research: heterophily and over-smoothing, arguing that both stem from the static, low-pass filter design of models like ChebyNet.

Instead of proposing separate, complex architectural solutions for each problem, we \textbf{advanced a unified solution} at the filter level by proposing `LaguerreNet`. This novel filter, based on continuous Laguerre polynomials, extends the class of adaptive polynomial filters (like the discrete MeixnerNet) into the continuous domain. By making its $\alpha$ parameter learnable, `LaguerreNet` adapts its spectral shape to the underlying graph.

We demonstrated that a `LayerNorm`-based stabilization strategy successfully tames the numerical instability of `LaguerreNet`'s unbounded $O(k^2)$ recurrence coefficients.

Our experiments confirmed the power of this approach:
\begin{enumerate}
    \item \textbf{Heterophily:} The adaptive filter family (`LaguerreNet`, `MeixnerNet`) achieves SOTA results on heterophilic benchmarks, validating the adaptive approach.
    \item \textbf{Over-smoothing:} `LaguerreNet` is highly robust to over-smoothing. Unlike `ChebyNet` (which collapses) and `KrawtchoukNet` (stable by design), `LaguerreNet` achieves stability \textit{despite} its unbounded coefficients, allowing its performance to increase up to $K=10$.
\end{enumerate}

This work positions `LaguerreNet` as a powerful, simple, and efficient addition to the adaptive polynomial filter class, offering a strong alternative to complex IIR filters (ARMAConv) and architectural GNNs (GCNII), and opening new avenues for exploring the Askey scheme.



\begin{thebibliography}{99}
\itemsep 1pt

\bibitem{shuman2013emerging}
D. I. Shuman, S. K. Narang, P. Frossard, A. Ortega, and P. Vandergheynst, "The emerging field of signal processing on graphs," \textit{IEEE Signal Processing Magazine}, vol. 30, no. 3, pp. 83-98, 2013.

\bibitem{bruna2014spectral}
J. Bruna, W. Zaremba, A. Szlam, and Y. LeCun, "Spectral networks and locally connected networks on graphs," in \textit{Intl. Conf. on Learning Representations (ICLR)}, 2014.

\bibitem{defferrard2016convolutional}
M. Defferrard, X. Bresson, and P. Vandergheynst, "Convolutional neural networks on graphs with fast localized spectral filtering," in \textit{Advances in Neural Information Processing Systems (NIPS)}, 2016.

\bibitem{kipf2017semi}
T. N. Kipf and M. Welling, "Semi-supervised classification with graph convolutional networks," in \textit{Intl. Conf. on Learning Representations (ICLR)}, 2017.

\bibitem{zhu2020beyond}
J. Zhu, Y. Wang, H. Wang, J. Zhu, and J. Tang, "Beyond homophily in graph neural networks: Current limitations and open challenges," in \textit{Proc. ACM SIGKDD Intl. Conf. on Knowledge Discovery \& Data Mining (KDD)}, 2020.

\bibitem{li2018deeper}
Q. Li, Z. Han, and X. Wu, "Deeper insights into graph convolutional networks for semi-supervised learning," in \textit{AAAI Conf. on Artificial Intelligence}, 2018.

\bibitem{askey1985beta}
R. Askey and J. Wilson, "Some basic hypergeometric orthogonal polynomials that generalize Jacobi polynomials," \textit{Memoirs of the American Mathematical Society}, vol. 54, no. 319, 1985.

\bibitem{goksu2025meixnernet}
H. Göksu, "MeixnerNet: Adaptive and robust spectral graph neural networks with discrete orthogonal polynomials," \textit{IEEE Signal Processing Letters}, 2025. (Submitted for review).

\bibitem{ba2016layer}
J. L. Ba, J. R. Kiros, and G. E. Hinton, "Layer normalization," \textit{arXiv preprint arXiv:1607.06450}, 2016.

\bibitem{he2021bernnet}
M. He, Z. Wei, and H. Huang, "BernNet: Learning arbitrary graph spectral filters via Bernstein polynomials," in \textit{Advances in Neural Information Processing Systems (NeurIPS)}, 2021.

\bibitem{levie2018cayleynets}
R. Levie, F. Monti, X. Bresson, and M. M. Bronstein, "CayleyNets: Graph convolutional neural networks with complex rational spectral filters," \textit{IEEE Transactions on Signal Processing}, vol. 67, no. 1, pp. 97-112, 2018.

\bibitem{bianchi2021graph}
F. M. Bianchi, D. Grattarola, C. Alippi, and L. Livi, "Graph neural networks with convolutional ARMA filters," \textit{IEEE Transactions on Pattern Analysis and Machine Intelligence (TPAMI)}, vol. 45, no. 5, pp. 5999-6011, 2021.

\bibitem{isufi2024graph}
E. Isufi, A. G. Marques, D. I. Shuman, and S. Segarra, "Graph filters for signal processing and machine learning on graphs," \textit{IEEE Signal Processing Magazine}, vol. 41, no. 2, pp. 12-32, 2024.

\bibitem{li2025ergnn}
G. Li, J. Yang, and S. Liang, "ERGNN: Spectral graph neural network with explicitly-optimized rational graph filters," in \textit{IEEE Intl. Conf. on Acoustics, Speech and Signal Processing (ICASSP)}, 2025.

\bibitem{abu2019mixhop}
S. Abu-El-Haija, B. Perozzi, A. Kapoor, N. Alipourfard, K. Lerman, H. Harutyunyan, and G. Ver Steeg, "MixHop: Higher-order graph convolutional architectures via sparse matrix powering," in \textit{Intl. Conf. on Machine Learning (ICML)}, 2019.

\bibitem{pei2020geomgcn}
H. Pei, B. Wei, K. C. C. Chang, Y. Lei, and B. Yang, "Geom-GCN: Geometric graph convolutional networks," in \textit{Intl. Conf. on Learning Representations (ICLR)}, 2020.

\bibitem{velickovic2018graph}
P. Veličković, G. Cucurull, A. Casanova, A. Romero, P. Liò, and Y. Bengio, "Graph attention networks (GAT)," in \textit{Intl. Conf. on Learning Representations (ICLR)}, 2018.

\bibitem{bo2021beyond}
D. Bo, X. Wang, C. Shi, H. Shen, "Beyond low-frequency information in graph convolutional networks," in \textit{Proc. AAAI Conf. on Artificial Intelligence}, 2021.

\bibitem{zhu2021graph}
J. Zhu, R. R. L. Leavell, L. M. Kaplan, S. T. Chowdhury, and E. B. Khalil, "Graph neural networks with heterophily (CPGNN)," in \textit{Proc. AAAI Conf. on Artificial Intelligence}, 2021.

\bibitem{gasteiger2019predict}
J. Gasteiger, A. Bojchevski, and S. Günnemann, "Predict then propagate: Graph neural networks meet personalized pagerank," in \textit{Intl. Conf. on Learning Representations (ICLR)}, 2019.

\bibitem{chien2021adaptive}
E. Chien, J. Liao, W. H. Chang, and C. K. Yang, "Adaptive graph convolutional neural networks (GPR-GNN)," in \textit{Intl. Conf. on Learning Representations (ICLR)}, 2021.

\bibitem{xu2018representation}
K. Xu, C. Li, Y. Tian, T. Sonobe, K. Kawarabayashi, and S. Jegelka, "Representation learning on graphs with jumping knowledge networks," in \textit{Intl. Conf. on Machine Learning (ICML)}, 2018.

\bibitem{chen2020simple}
M. Chen, Z. Wei, Z. Huang, B. Ding, and Y. Li, "Simple and deep graph convolutional networks," in \textit{Intl. Conf. on Machine Learning (ICML)}, 2020.

\bibitem{goksu2025krawtchouknet}
H. Göksu, "KrawtchoukNet: Solving GNN over-smoothing with numerically stable discrete global spectral filters," \textit{IEEE Transactions on Neural Networks and Learning Systems}, 2025. (Submitted for review).

\bibitem{goksu2025charliernet}
H. Göksu, "CharlierNet: A minimalist and competitive spectral GNN filter for resource-constrained sensor networks," \textit{IEEE Sensors Letters}, 2025. (Submitted for review).

\bibitem{sen2008collective}
P. Sen, G. Namata, M. Bilgic, L. Getoor, B. Galligher, and T. Eliassi-Rad, "Collective classification in network data," \textit{AI Magazine}, vol. 29, no. 3, p. 93, 2008.

\bibitem{yang2016revisiting}
Z. Yang, W. W. Cohen, and R. Salakhutdinov, "Revisiting semi-supervised learning with graph embeddings," in \textit{Intl. Conf. on Machine Learning (ICML)}, 2016.

\bibitem{hamilton2017inductive}
W. L. Hamilton, R. Ying, and J. Leskovec, "Inductive representation learning on large graphs (GraphSAGE)," in \textit{Advances in Neural Information Processing Systems (NeurIPS)}, 2017.
\end{thebibliography}
\end{document}